\documentclass[10pt,twocolumn,letterpaper]{article}

\usepackage{iccv}
\usepackage{times}
\usepackage{epsfig}
\usepackage{graphicx}
\usepackage{amsmath}
\usepackage{amssymb}
\usepackage[linesnumbered,ruled,vlined]{algorithm2e}
\usepackage{makecell}
\usepackage{mathtools}

\DeclarePairedDelimiter\floor{\lfloor}{\rfloor}
\setcellgapes{1.5pt}
\usepackage[pagebackref=true,breaklinks=true,letterpaper=true,colorlinks,bookmarks=false]{hyperref}
\usepackage[accsupp]{axessibility}

\iccvfinalcopy % *** Uncomment this line for the final submission

\def\iccvPaperID{****} % *** Enter the ICCV Paper ID here

\ificcvfinal\fi

\begin{document}

%%%%%%%%% TITLE
\title{HIRE-SNN: \underline{H}arnessing the \underline{I}nherent \underline{R}obustness of \underline{E}nergy-Efficient\\
Deep \underline{S}piking \underline{N}eural \underline{N}etworks by Training with Crafted Input Noise}

\author{Souvik Kundu, Massoud Pedram, Peter A. Beerel\\
University of Southern California,
Los Angeles, CA, USA\\
{\tt\small \{souvikku, pedram, pabeerel\}@usc.edu}
%\and
}

\maketitle
\ificcvfinal\thispagestyle{empty}\fi

%%%%%%%%% ABSTRACT
\begin{abstract}
Low-latency deep spiking neural networks (SNNs) have become a promising alternative to conventional artificial neural networks (ANNs) because of their potential for increased energy efficiency on event-driven neuromorphic hardware.% Low-latency SNNs have potential computational benefits compared to conventional ANNs. 
 Neural networks, including SNNs, however, are subject to various adversarial attacks and must be trained to remain resilient against such attacks for many applications. Nevertheless, due to prohibitively high training costs associated with SNNs, an analysis and optimization of deep SNNs under various adversarial attacks have been largely overlooked. In this paper, we first present a detailed analysis of the inherent robustness of low-latency SNNs against popular gradient-based attacks, namely fast gradient sign method (FGSM) and projected gradient descent (PGD). Motivated by this analysis, to harness the model's robustness against these attacks we present an SNN training algorithm that uses crafted input noise and incurs no additional training time. To evaluate the merits of our algorithm, we conducted extensive experiments with variants of VGG and ResNet on both CIFAR-10 and CIFAR-100 dataset. Compared to standard trained direct-input SNNs, our trained models yield improved classification accuracy of up to $13.7\%$ and $10.1\%$ on FGSM and PGD attack generated images, respectively, with negligible loss in clean image accuracy.
%through state-of-the-art direct-input hybrid training. 
Our models also outperform inherently-robust SNNs trained on rate-coded inputs with improved or similar classification performance on attack-generated images while having up to $25\times$ and $\mathord{\sim}4.6\times$ lower latency and computation energy, respectively.     
\end{abstract}
%%%%
%%%%
%%%%%%%%% BODY TEXT
\section{Introduction}
\label{sec:intro}
Artificial neural networks (ANNs) have become enormously successful in various computer vision applications \cite{ szegedy2015going, he2016deep, redmon2017yolo9000, tao2018image, kundu2021attentionlite}. However, as these applications are often part of safety-critical and trusted systems, concerns about their vulnerability to adversarial attacks have grown rapidly. In particular, well crafted adversarial images with small, often unnoticeable perturbations can fool a well trained ANN to make incorrect and possibly dangerous decisions \cite{moosavi2016deepfool, athalye2018obfuscated, xie2017adversarial}, despite their otherwise impressive performance on clean images. To improve the model performance of ANNs against attacks, training with various adversarially generated images \cite{madry2017towards, kundu2021dnr} has proven to be very effective. Few other prior art references \cite{zhang2019you, shafahi2019adversarial} have applied noisy inputs to train robust models. However, all these training schemes incur non-negligible clean image accuracy drop and require significant additional training time.

\begin{figure}[!t]
\includegraphics[width=0.46\textwidth]{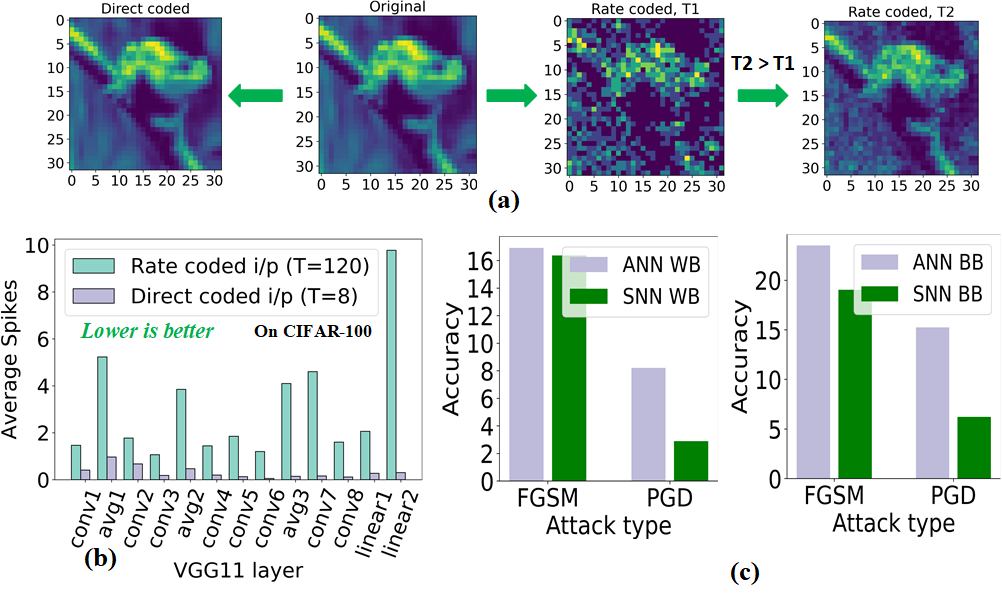}
\centering
   \caption{(a) Direct and rate-coded input variants of the original image. (b) Layer wise average spikes for VGG11. (c) Performance of direct-input VGG11 SNN and its equivalent ANN under various white-box (WB) and black-box (BB) attacks on CIFAR-100.}
   %over $T$ time steps.}
\label{fig:vgg11_sa_and_attack_perform}
\vspace{-6mm}
\end{figure}

Brain-inspired \cite{mainen1995reliability} deep spiking neural networks (SNNs)  
have also gained significant traction due to their potential for lowering the required power consumption of machine learning applications \cite{neuro_frontiers,spike_ratecoding}. The underlying SNN hardware can use binary spike-based sparse processing via accumulate (AC) operations over a fixed number of time steps\footnote{Here, a time step is the unit of time taken by each input image to be processed through all layers of the model.} $T$ which consume much lower power than the traditional energy-hungry multiply-accumulate (MAC) operations that dominate ANNs \cite{farabet2012comparison}. Recent advances in SNN training by using approximate gradient \cite{bellec_2018long} and hybrid direct-input-coded  ANN-SNN training with joint threshold, leak, and weight optimization \cite{rathi2020diet} have improved the SNN accuracy while simultaneously reducing the number of required time steps. This has lowered  both their computation cost, which is reflected in their average spike count as shown in Fig. \ref{fig:vgg11_sa_and_attack_perform}(b), and inference latency. However, the trustworthiness of these state-of-the-art (SOTA) SNNs under various adversarial attacks is yet to be fully explored. 

Some earlier works have claimed that SNNs may have $\textit{inherent}$ robustness against popular gradient-based adversarial attacks \cite{el2020securing, sharmin2020inherent, marchisio2020spiking}.
%This is especially useful for SNNs because training with adversarial images typically involves iterate gradient calculations over multiple time steps.
In particular, $\textit{Sharmin et al.} $ \cite{sharmin2020inherent} observed that rate-coded input-driven (Fig. \ref{fig:vgg11_sa_and_attack_perform}(a)) SNNs %with high $T$ 
have inherent robustness, which the authors primarily attributed to the highly sparse spiking activity of the model.
However, these explorations are mostly limited to small datasets on shallow SNN models, and more importantly, these techniques give rise to high inference latency.
This paper extends this analysis, asking two key questions. 
%
%addresses the question of whether these observations extend to deep SNN models as well as on direct-coded SNNs that have lower latency and relatively high spike activity. 
%Thus, we believe careful exploration of the SNN model performance under various attack scenario is necessary, particularly for the investigation of the following questions:
%
%This in turn reduces the layer activation sparsity and vastly limits the scope to leverage robustness advantage of deep SNNs via long inference time steps. 

1. \textit{To what degree does SOTA low-latency deep SNNs retain their inherent robustness under both black-box and white-box adversarial-attack generated images?}

2. \textit{Can computationally-efficient training algorithms improve the robustness of low-latency deep SNNs while retaining their high clean-image classification accuracy?}

\textbf{Our contributions are two-fold.} We first empirically study and provide detailed observations on inherent robustness claims about deep SNN models when the SNN inputs are directly coded. Interestingly, we observe that despite significant reductions in the average spike count, deep direct-input SNNs have lower classification accuracy compared to their ANN counterparts on  various white-box and black-box attack generated adversarial images, as exemplified in Fig. \ref{fig:vgg11_sa_and_attack_perform}(c). 
%of these low-latency SNNs as opposed to their high-latency counter part.
%. We provide insights in the context of SNN classification performances at reduced time steps.

Based on these observations, we present HIRE-SNN, a spike timing dependent backpropagation (STDB) based SNN training algorithm to better harness SNN's inherent robustness. In particular, we optimize the model trainable parameters using images whose pixel values are perturbed using crafted noise across the time steps. More precisely, we partition the training time steps $T$ into $\mathcal{N}$ equal-length periods of length $\floor{T/\mathcal{N}}$ and train each image-batch over each period, adding input noise after each period. 
The key feature of our approach is that, instead of showing the same image repeatedly, we efficiently use the time steps of SNN training to input different noisy variants of the same image. This avoids extra training time and, because we update the weights after each period, requires less memory for the storage of intermediate gradients compared to traditional SNN training methods.  
%with gradually increased noise. 
%We adapt the leak and threshold parameters concurrently with the weights, tuning them to
%to further make the distinctive behavioral parameters aware of 
%the noise-induced perturbations. 
%We show this approach harnesses the SNN model's inherent robustness without requiring additional training on adversarially-generated images.
%time cost. 
%This is particularly important as traditional SNN training is already expensive in terms of run-time and gradient storage 
%due its already expensive time and memory footprint caused by the 
%because it must iterate
%notion of iterative gradient storage 
%over multiple time steps. 
To demonstrate the efficacy of our scheme we conduct extensive evaluations with both VGG \cite{simonyan2014very} and ResNet \cite{he2016deep} SNN model variants on both CIFAR-10 and CIFAR-100 \cite{krizhevsky2009learning} datasets.  

The remainder of this paper is arranged as follows. In Section \ref{sec:back} and \ref{sec:snn_inherent} we present the necessary background and provide analysis of inherent robustness of direct-input SNNs, respectively. Section \ref{sec:scin} presents  our training scheme. We provide our experimental results and discussion on Section \ref{sec:expt} and finally conclude in Section \ref{sec:conc}. 
%%%%
%%%%
\section{Background}
\label{sec:back}
%%%%
%%%%
\begin{figure}[!t]
\includegraphics[width=0.48\textwidth]{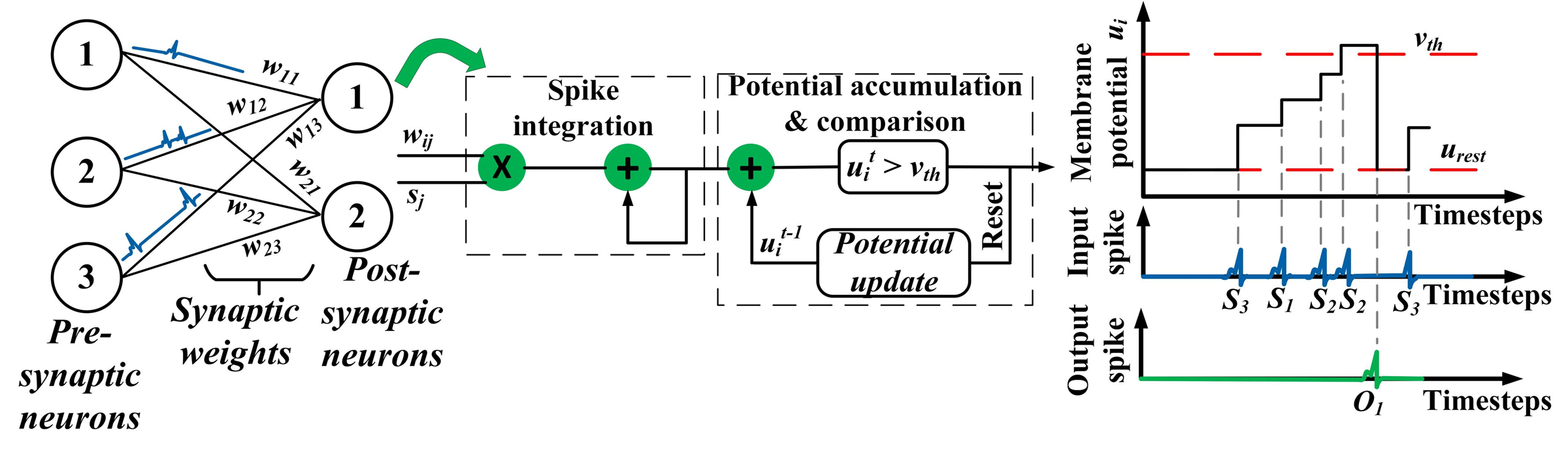}
\centering
   \caption{SNN fundamental operations.}
\label{fig:snn_fundamental}
\vspace{-4mm}
\end{figure}
%%%%
%%%%
\subsection{SNN Fundamentals}
\label{subsec:snn_funda}

In ANN training, updating weights involve a single forward-backward pass transferring multi-bit weights and gradients through layers of a network. In contrast, in SNN training, updating weights involve $T$ forward and backward passes, each pass propagating either binary spikes or a notion of their gradients. 
Note that $T$ is known as the SNN's inference latency 
and the spiking dynamics of an SNN layer are typically defined with either the Integrate-Fire (IF) \cite{lu2020exploring} or Leaky-Integrate-Fire (LIF) \cite{leefin2020} neuron model. Interestingly, the LIF model introduces a non-linearity in the model that can be compared to the rectilinear (ReLU) operation in conventional ANNs.  The discrete time \cite{wu2018spatio} iterative version of the LIF neuron dynamics is defined by the following equation
%%%%
%%%%
\begin{align}
    u_i^{t+1} &= \lambda u_i^t + \sum_{j}w_{ij}O_j^{t} - v_{t}O_i^t\\
    O_i^{t}   &=
    \begin{cases}
    1, & \text{if } z_i^t>0\\
    0, & \text{otherwise}
    \end{cases}
\label{eq:neuron_discrete}
\end{align}
\noindent
where $z_i^{t}$ = ($\frac{u_i^t}{v_{t}} -1$) is the normalized membrane potential and $v_{t}$ is current layer firing threshold. The decay factor $\lambda$=1 for IF and $\lambda < 1$ for LIF neuron models. $u_i^{t+1}$ represents the membrane potential of the $i^{th}$ neuron at time step $t+1$, and $O_i^{t}$ and $O_j^{t}$ represent the output spikes of current neuron $i$ and one of its pre-synaptic neurons $j$, respectively. $w_{ij}$ represents the weight between the two. The inference output is obtained by comparing the total number of spikes generated by each output neuron over $T$ time steps. 

However, supervised training of SNNs faces the challenge of backpropagating gradients of binary spike trains which are undefined. This issue has been addressed using approximate gradient computations \cite{bellec_2018long} at the cost of slow convergence and high memory requirements. In this paper,
we refer to this step as $\textit{traditional}$ SNN training.

\textbf{ANN-to-SNN conversion.} 

A popular alternative to traditional SNN training {\em from scratch} for deep SNNs involves first training a constrained ANN model \cite{dsnn_conversion_abhronilfin} and then converting it into an SNN by computing layer thresholds \cite{dsnn_conversion_ijcnn, dsnn_conversion_abhronilfin}. 
The SNN models yielded through this technique, however, require high latency $T$ to perform well on complex vision tasks. We thus adopt a more recently developed hybrid training technique that leverages the benefits of the ANN-to-SNN conversion technique followed by a few epochs of direct-input driven traditional SNN training to reduce inference latency \cite{rathi2020diet}. 

\subsection{Adversarial Attacks}
\label{subsec:attack}
Various gradient-based adversarial attacks have 
been proposed to generate adversarial images, which have barely-visible perturbations from the original images but still manage to fool a trained neural network. One such attack is the fast gradient sign method (FGSM) \cite{goodfellow2014explaining}. Let $f(\textit{\textbf{x}}, \textit{\textbf{y}})$ represents the function of an ANN, implicitly parameterized by network parameters $\boldsymbol{\theta}$, that accepts a vectorized input image \textit{\textbf{x}} and generates a corresponding label \textit{\textbf{y}}. FGSM perturbs each element \textit{x} in \textit{\textbf{x}} along the sign of %($sgn(.)$) 
the gradient of the inference loss w.r.t. \textit{\textbf{x}} 
\begin{align}
\vspace{-0.5cm}
\hat{\textbf{\em x}}&={\textbf{\em x}} + \epsilon * sign(\nabla_{x}\mathcal{L}(f(\textbf{\em  x}, \textbf{\em  y}))) \label{eq:fgsm} 
\vspace{-0.5cm}
\end{align}
\noindent
where the scalar $\epsilon$ is the perturbation parameter that determines the severity of the attack.

Another well-known attack is projected gradient descent (PGD) \cite{madry2017towards}. It is a multi-step variant of FGSM and is known to be one of the most powerful first-order attacks \cite{athalye2018obfuscated}. 
Assuming $\hat{\textbf{\em x}}^{k=1} = {\textbf{\em x}}$ the iterative update of the perturbed data $\hat{\textbf{\em x}}$ in $k^{th}$ step of PGD is given in Eq. \ref{eq:pgd}.   
\begin{align}
\vspace{-0.5cm}
\hat{\textbf{\em x}}^{k}&=\textit{Proj}_{P_{\epsilon}(\textit{\textbf{x}})} [\hat{\textbf{\em x}}^{k-1} + \alpha * sign(\nabla_{x}\mathcal{L}(f(\hat{\textbf{\em x}}^{k-1}, \textbf{ \em y})))] \label{eq:pgd}
\vspace{-0.5cm}
\end{align}
\noindent
Here, $\textit{Proj}$ projects the updated adversarial sample onto the projection space $P_{\epsilon}(\textit{\textbf{x}})$, the   $\epsilon$-$L_{\infty}$ neighbourhood of the benign sample\footnote{It is noteworthy that the generated $\hat{\textbf{\em x}}$ are clipped to a valid range which for our experiments is $[0,1]$.} $\textit{\textbf{x}}$, and $\alpha$ is the attack step size. 

Note that for both attack techniques we consider two scenarios: 1) white-box (WB) attack in which the attacker has complete access to the model parameters, and 2) black-box (BB) attack in which the attacker has no knowledge of the model's trainable parameters and thus produces weaker perturbations than the white-box alternative. % For comprehensive analysis we consider both scenarios. 
%for PGD. 

\begin{table*}
\scriptsize\addtolength{\tabcolsep}{-0.5pt}
\begin{center}
{\makegapedcells
\begin{tabular}{c|c|c|c||c|c|c||c|c|c}
\hline
{} & \multicolumn{3}{|c||}{Accuracy (\%) with} & \multicolumn{3}{|c||}{Accuracy (\%) with} & \multicolumn{3}{|c}{Accuracy (\%) with} \\
Model- & \multicolumn{3}{|c||}{ANN} & \multicolumn{3}{|c||}{high latency SNN-BP \cite{sharmin2020inherent}} & \multicolumn{3}{|c}{low latency SNN-BP } \\
\cline{2-10}
{Attack category} & Clean & FGSM & PGD & Clean & FGSM & PGD & Clean  & FGSM & PGD \\
\hline
 \multicolumn{10}{c}{Dataset : CIFAR-10} \\
\hline
VGG5-WB & 90.2 & 13.3 & 2.0 & $\textit{89.3}$ & $\textit{15.0}$ & $\textit{3.8}$ & 87.9 & 35.5 & 5.3 \\
%\cline{2-8}
VGG5-BB & 90.2 & 24.0 & 6.4 & $\textit{89.3}$ & $\textit{21.5}$ & $\textit{16}$ & 87.9 & 38.3 & 6.7 \\
\hline
ResNet12-WB & 92.6 & 19.9 & 2.0 & -- & -- & -- & 91.9 & 21.1 & 0.2 \\
%\cline{2-8}
ResNet12-BB & 92.6 & 28.6 & 4.3 & -- & -- & -- & 91.9 & 24.7 & 0.6 \\
\hline
 \multicolumn{10}{c}{Dataset : CIFAR-100} \\
\hline
VGG11-WB & 69.5 & 16.9 & 8.2 &  $\textit{64.4}$ & $\textit{15.5}$ & $\textit{6.3}$ & 65.6 & 16.4 & 2.9\\
%\cline{2-8}
VGG11-BB & 69.5 & 23.5 & 15.3 & $\textit{64.4}$ & $\textit{21.4}$ & $\textit{16.5}$ & 65.6 & 19.0 & 6.2\\
\hline
ResNet12-WB & 61.5 & 13.5 & 2.8 &  -- & -- & -- & 61.9 & 10.5 & 0.6 \\
%\cline{2-8}
ResNet12-BB & 61.5 & 23.2 & 12.0 & -- & -- & -- & 61.9 & 14.1 & 2.0\\
\hline
\end{tabular}}
\end{center}
\caption{Comparison of model performances under various white-box and black-box attacks on both CIFAR-10 and CIFAR-100. Note that \textit{italicized} values are taken directly from the original paper.}
\vspace{-4mm}
\label{tab:comparsion_table_with_normal_train}
\end{table*}

\section{Initial Study: SNN Robustness Analysis}
\label{sec:snn_inherent}

To motivate our novel training algorithm to harness robustness, this section describes an empirical analysis into the robustness of traditionally-trained SNNs on gradient-based adversarial attacks. We performed traditional SNN training with the initial weights and thresholds set to that of a trained equivalent ANN and generated through the conversion process, respectively\footnote{The description of our training hyperparameter settings is given in Section \ref{subsec:setup} for all the experiments in this section.}.  

\subsection{Performance Analysis}

We first performed SNN training with direct-coded inputs and evaluated the robustness of the trained models under various white-box and black-box attacks. Interestingly, as shown in Table \ref{tab:comparsion_table_with_normal_train}, the generated deep SNNs, i.e., VGG11 and ResNet12, consistently provide inferior performance against various black-box attacks compared to their ANN counterparts. 
For example, we observe that the VGG11 SNN provides only $6.2\%$ accuracy on the PGD black-box attack, while its ANN equivalent provides an accuracy of $15.25\%$. These results imply that traditional SNN training appears to be insufficient to harness the inherent robustness of low-latency direct-input deep SNNs. 

It is important to note that \cite{sharmin2020inherent} observed that, for rate-coded SNNs, spike based sparse activation maps correlates with adversarial success. To extend this analysis to direct-input SNNs, we examine two distinct metrics of the SNN's spiking activity, as defined below. 

\textbf{Definition 1.} \textit{Spiking activity (SA)}: We define a layer's spiking activity as the ratio of number of spikes produced over all the neurons accumulated across all time units $T$ of a layer to the total number of neurons present in that layer. We also define a layer's SA divided by $T$ as the \textit{time averaged spiking activity} (TASA).

\begin{figure}[!t]
\includegraphics[width=0.42\textwidth]{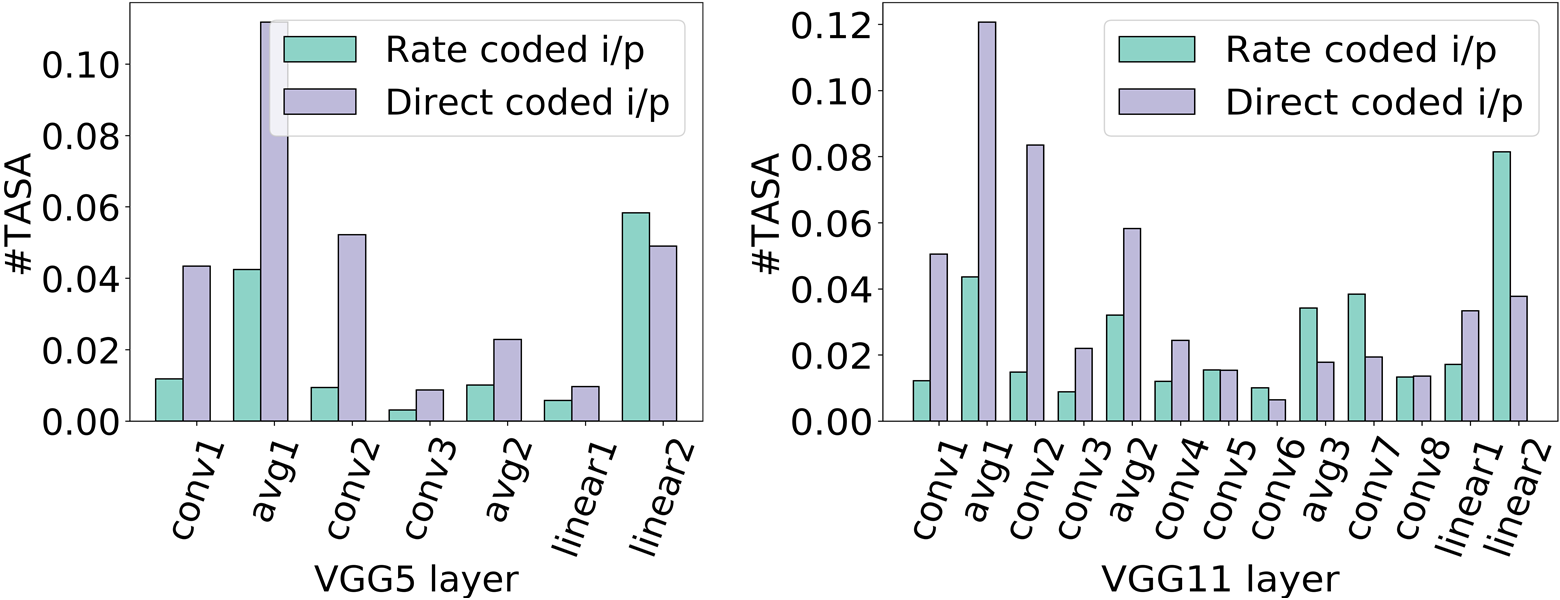}
\centering
   \caption{Per layer TASAs of VGG5 and VGG11 on CIFAR-10 and CIFAR-100, respectively.}
\label{fig:vgg11_vgg5_tasa}
\vspace{-2mm}
\end{figure}

\textbf{Observation 1.} \textit{Compared to rate-coded SNNs, deep SNNs with direct-coded inputs and lower latency generally exhibit lower SA but higher TASA, especially in the initial convolution layers.}

%Fig. \ref{fig:vgg11_vgg5_tasa} shows that, compared to their rate-coded counterparts, the TASA of direct-coded VGG5 and VGG11 SNNs increases 
%This means that 
%despite a significant decrease in overall spiking activity (Fig. \ref{fig:vgg11_sa_and_attack_perform}), particularly, at the initial layers.

 Particularly, this distinction can be seen for VGG11 SNN 
in Figs. \ref{fig:vgg11_vgg5_tasa}(b) and \ref{fig:vgg11_sa_and_attack_perform}(b)
%
%, inference in low-latency direct-coded SNNs involves more spikes per time step, especially at the initial layers.
and suggests that the observation that sparse SA correlates well with success against adversarial images \cite{sharmin2020inherent} can extend to low-latency direct-coded SNNs if the spiking activity is quantified using TASA. VGG5 also shows lower sparsity level of spikes at the early layers (Fig. \ref{fig:vgg11_vgg5_tasa}(a)) compared to its rate-coded counterpart.

%Thus, compared to SA, we believe TASA is a better measure of spiking activity as it relates to the degree of adversarial success \cite{sharmin2020inherent}.
%a more notion of layer activation sparsity and understand that an increased TASA means reduction in activation sparsity. We hypothesize this phenomena to play a pivotal role in reducing the SNN model performance particularly on strong black-box attack generated images through PGD, as increased activation sparsity generally benefits adversarial success \cite{sharmin2020inherent}.

\begin{figure}[!tb]
\includegraphics[width=0.40\textwidth]{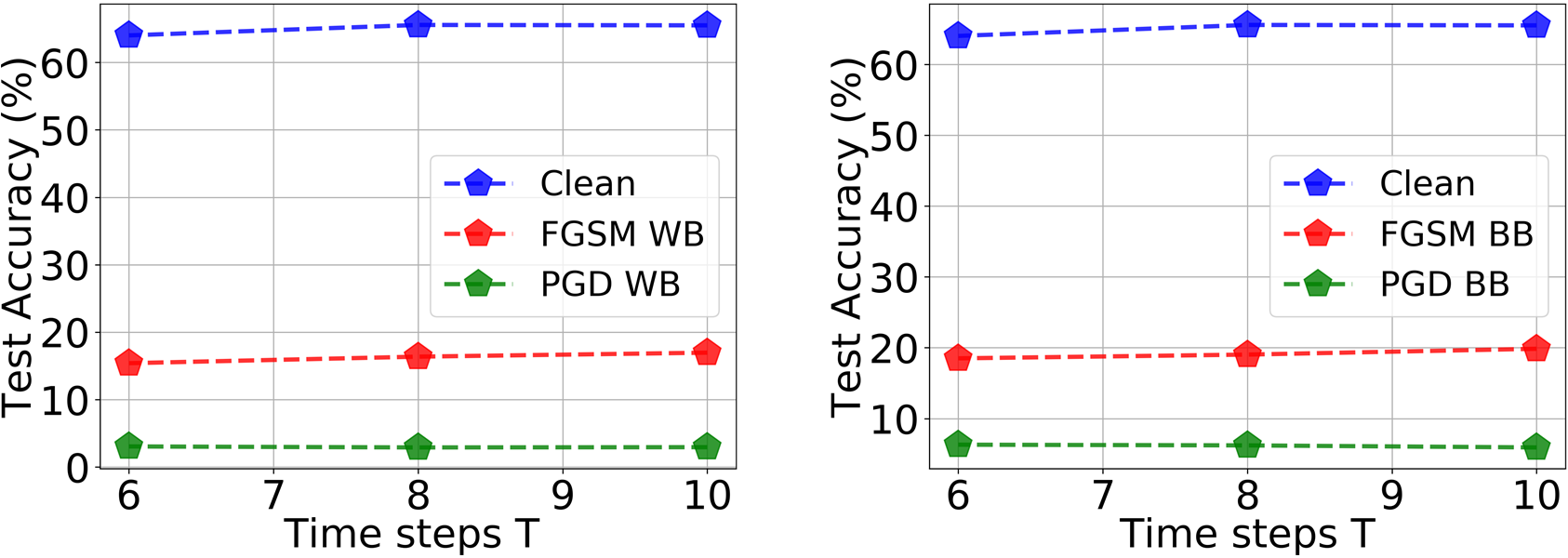}
\centering
   \caption{Classification performance of VGG11 on CIFAR-100 as number of time steps $T$ varies.}
\label{fig:vgg11_c100_tstep_vs_acc}
\vspace{-2mm}
\end{figure}

\textbf{Observation 2.} \textit{Direct-input coded SNNs yield lower clean accuracy and no significant improvement in adversarial image classification accuracy as latency $T$ is reduced.}

Earlier research has shown that robustness to adversarial images of SNNs trained on rate-coded inputs improves with the reduction in training time steps %while causing insignificant drop in clean image accuracy
\cite{sharmin2020inherent}. Motivated by this we performed a similar analysis on VGG11 using 
direct input CIFAR-100. Interestingly, as shown in Fig. \ref{fig:vgg11_c100_tstep_vs_acc}, as $T$ reduces the classification performance on both black box and white box attack generated images does not improve. Intuitively, these attacks are more effective on direct-coded inputs because of the lack of approximation at the inputs, unlike for Poisson generated rate-coded inputs.
%that yield inherent robustness \cite{sharmin2020inherent}. 
However, SNNs with rate-coded inputs generally require larger training time and memory footprint \cite{rathi2020diet} to reach competitive accuracy.
%when fed with such spike based approximate inputs. 
In Fig. \ref{fig:vgg11_c100_tstep_vs_acc} we relate the reduction in network performance on clean images to the aggressive reduction in the number of training time steps.  

\textbf{Definition 2.} \textit{Perturbation distance (PD):} We define perturbation distance as the $L_2$-norm of the absolute difference of pixel values between a real image and its adversarially-perturbed variant. Similarly, for an intermediate layer with spike based activation maps \cite{Kundu2021WACV, kundu2021towards}, we define $\textit{spike PD}$ as the $L_2$-norm of the absolute difference of the normalized spike-based activation maps generated at a layer output when fed with an original and its perturbed variant, respectively. 
\begin{figure}[!ht]
\includegraphics[width=0.40\textwidth]{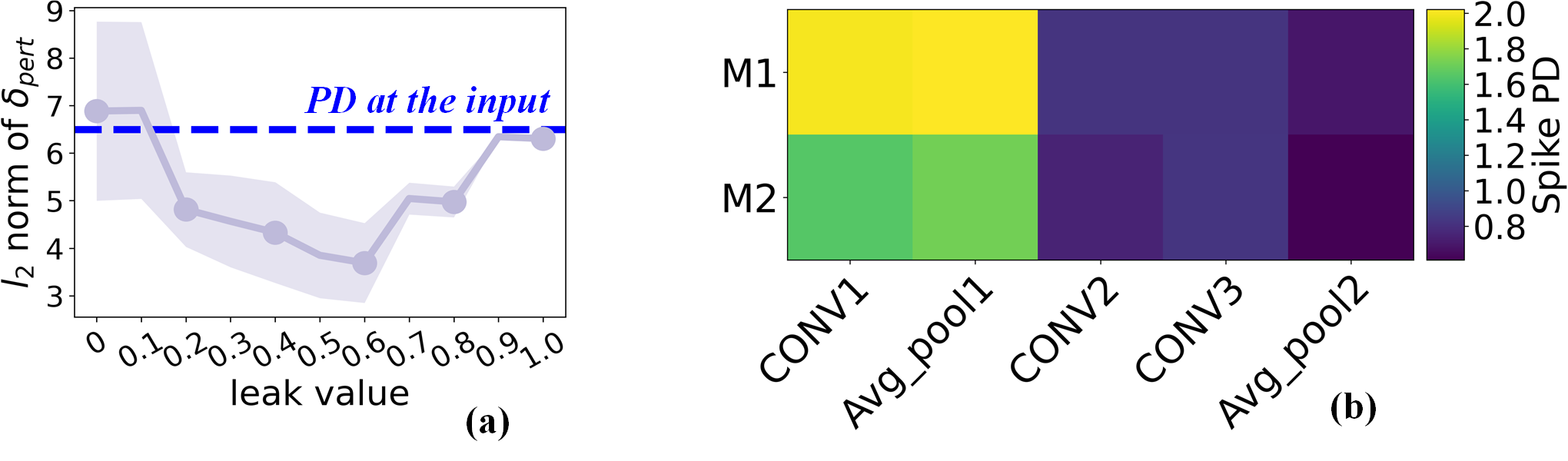}
\centering
   \caption{(a) PD vs LIF leak parameter for a fixed threshold (0.8) and latency ($T=10$) averaged over two randomly chosen input images that are perturbed with PGD-1. (b) Intermediate layer spike PD for VGG5 fed with a randomly-selected CIFAR-10 clean image and its perturbed variant.}
\label{fig:leak_vs_perturb_at_inp}
\vspace{-2mm}
\end{figure}

\textbf{Observation 3.} \textit{Leaky integrate and fire (LIF) non-linearity applying layers contribute to the inherent robustness of rate-coded input driven SNNs by diminishing the perturbation distance \cite{sharmin2020inherent}. 
Unfortunately, this observation does not generally hold for direct-coded SNNs in which the LIF layers may increase or degrade the perturbation distance, suggesting that the impact of the leak parameter must be considered jointly with other factors, including related weights and thresholds.}

The LIF operation in SNNs yields non-linear dynamics that can be contrasted to the piecewise linear ReLU operation in traditional ANNs. To analyze their impact on image perturbation distance, we fed an LIF layer the clean images taken from a digit classification dataset \cite{deng2012mnist}  along with their perturbed variants, sweeping the leak parameter value and measuring the impact on the perturbation distance. As depicted in Fig. \ref{fig:leak_vs_perturb_at_inp}(a), the leak factor helps reduce the perturbation distance only if its value falls in a certain range. 

To further study the impact of LIF layers, we analyzed the spike PD of the models. In particular, we fed two VGG5 SNN models trained with two different seeds ($M1$ and $M2$) with a randomly-sampled CIFAR-10 clean image $\textbf{\em {x}}_C$ and its black-box attack generated variant $\textbf{\em {x}}_P$ and computed the corresponding intermediate layer spike PDs. Both $M1$ and $M2$ classified $\textbf{\em {x}}_C$ correctly, however, $M2$ failed to correctly classify $\textbf{\em {x}}_P$. Interestingly, as shown in Fig. \ref{fig:leak_vs_perturb_at_inp}(b) despite the presence of the LIF layers,  the spike PD values do not always reduce as we progress from layer to layer through the network. %move from layer $l-1$ to layer $l$. 
Moreover, this degree of unpredictability seems to be irrespective of whether the model classifies the image correctly. We conclude that despite LIF's promise to reduce input perturbation, its impact is also a function of other parameters, including the trainable weights, leak, threshold, and time steps. 
%Thus we believe more careful consideration of these parameters are required before certifying LIF to be beneficial against adversarial perturbation.   

Based on these empirical observations, we assert that the majority of the reasons that make rate-coded SNN inherently robust are either absent or need careful tuning for direct-input SNN models, as presented in the next section.  
%%%%
%%%%
\begin{figure}[!t]
\includegraphics[width=0.38\textwidth]{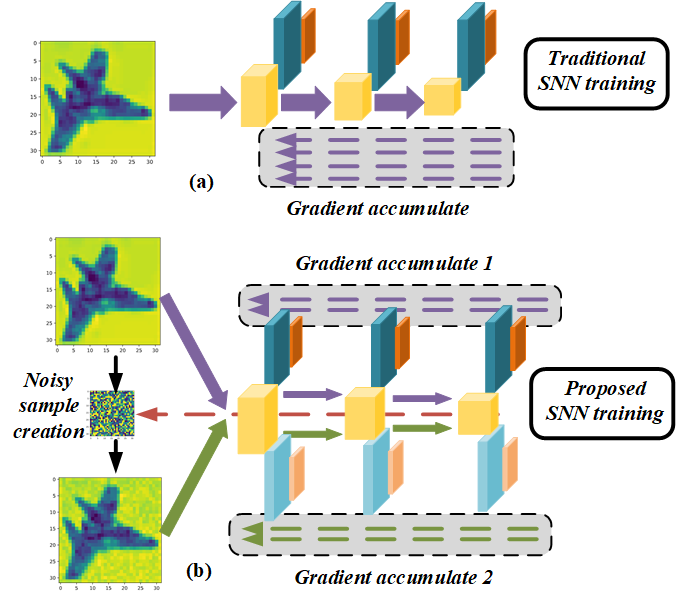}
\centering
   \caption{(a) and (b) represent the traditional and proposed training schemes, respectively. Here the green and orange blocks represent activation maps and the gradients that are generated after passing the input image. For the proposed training scheme we use two color variants deep and light, respectively, to highlight the sets of activation maps and gradients from an image and its noisy variant during two different periods. The yellow blocks represent the weight tensors that get updated from accumulated gradients. In (b) we compute the input gradient with these updated weights to craft the noise. Here, we assumed $T = 4$ and $\mathcal{N} = 2$.}
   %over $T$ time steps.}
\label{fig:SCIN_training}
\vspace{-4mm}
\end{figure}
%%%%
%%%%
\section{HIRE-SNN Training}
\label{sec:scin}

This section presents our training algorithm for robust SNNs. %training technique. 
%that better harnesses the inherent robustness of SNNs compared to conventional techniques. %generated SNNs.  
As shown in Eq. \ref{eq:neuron_discrete} the LIF neuron functional output at each
time step recursively depends on its state in previous time steps 
%that introduces an implicit recurrent connection within the model
\cite{neftci_surg}. 
%Rate-coded inputs generated with the help of a Poisson random generator \cite{diehl2015fast} are typically used to provide spike-based inputs.
%for each time step.  
Each input pixel in traditional SNN training using direct-coded inputs, is fed into the network as a multi-bit value that is {\em fixed} over the $T$ time steps and yield an order of magnitude reduction in latency compared to rate-coded alternatives. 
%This helps reduce $T$ for both training and inference by an order of magnitude.
%$\mathord{\sim}10\times$. %However, all the earlier input coding techniques assumed the conventional frame-based input image does not change throughout the $T$ time steps.
%
However, our approach is different than direct coding because we partition the training time steps $T$ into $\mathcal{N}$ equal-length periods and feed in a different perturbed variant of the image during each period of $\floor{T/\mathcal{N}}$ steps. 

To be more precise, consider an SNN model defined by the function $g(\textit{\textbf{x}}, \textit{\textbf{y}}; T)$ implicitly parameterized by $\boldsymbol{\theta}$.
Assume an input batch $\mathcal{B}$ of size $h_i \times w_i \times c_i \times n_{\mathcal{B}}$, where $h_i, w_i$, and $c_i$ represent spatial height, width, and channel length of an image, respectively, with $n_{\mathcal{B}}$ as the number of images in the batch. %This batch in normal SNN training is fed to the SNN for $T$ steps with no changes in its pixel values. 
In contrast to traditional approaches, where weight update happens only after $T$ steps, we allow different perturbed image variants generation and weight update to happen at small interval of $\floor{T/\mathcal{N}}$ steps within the window of $T$, for an image batch. This important modification allows us to train the model with different adversarial image variants without costing any additional training time.
%We repeat this for all images in a batch until for all periods.
%More precisely, we start training with a perturbation tensor of size $h_i \times w_i \times c_i \times n_{\mathcal{B}}$ set to zero and, for the first $\floor{T/\mathcal{N}}$ steps, with the first batch of image we feed the original input to the model to compute the loss incurred during that period. 
As the exact gradient of the binary spike trains is undefined, we use a linear surrogate gradient (Eq. \ref{eq:linear_surrogate}) approximation \cite{bellec_2018long} to allow backpropagation (BP) and gradient-based parameter update in SNNs. 
\begin{align}
\vspace{-0.5cm}
    & \frac{\partial O_i^{t}}{\partial z_i^{t}} = \gamma * max\{0, 1-|z_i^{t}|\} 
    \label{eq:linear_surrogate} 
\vspace{-0.5cm}
\end{align}
where $\gamma$ is a damping factor that controls the approximate back-propagation error, to update the trainable parameters.
%of the model following standard gradient-based parameter update. 
We also compute the gradient of the loss with respect to each input pixel {\em x} to craft the perturbation for next period. Through an abuse of notation, we define $\epsilon_{s}$ and $\epsilon_{t}$ as the pixel noise step and bound, respectively, and generate perturbation scalar for each of the $h_i \times w_i \times c_i$ pixels of an image as 
\begin{align}
\vspace{-0.6cm}
 & \kappa = clip[\kappa + \epsilon_{s} \times sign(\nabla_{x}\mathcal{L}), -\epsilon_t, +\epsilon_t] \label{eq:perturb_train} 
\vspace{-0.6cm}
\end{align}
\noindent
 where $\kappa$ represents the perturbation for an input pixel $\textit{x}$ of a batch $\mathcal{B}$ computed at the $p^{th}$ period. 
 Note that for current batch, we initialize $\kappa$ in the first period with the perturbation computed at the last period of the previous batch. In contrast, the computation of the perturbation of other periods is based on the computed 
 perturbation from the corresponding previous period.
 %perturbed images with the perturbation generated on its own images. Also, the $p^{th}$ period computes the perturbation based on the updated weights of $(p-1)^{th}$ period. 
 It is noteworthy that $\epsilon_s$ is not necessarily the same as $\epsilon$ of the FGSM or PGD attacks, and we generally choose $\epsilon_s$ to be sufficiently small 
 %such that there is no significant perturbation on the original images for the network 
 to not lose significant classification accuracy on clean images.  
 We include weights $\textbf{W}$, threshold $\textbf{\textit{v}}_t$ and leak $\textbf{\textit{l}}_k$ parameters in the trainable parameters $\boldsymbol{\theta}$ to retain clean image accuracy at low latencies \cite{rathi2020diet}. Our detailed training algorithm called HIRE-SNN is presented in Algorithm \ref{alg:scin}. It is noteworthy that, apart from noise crafted inputs, our training framework can easily be extended to support various input encoding \cite{datta2021training, sharmin2020inherent} as well as image augmentation techniques \cite{rebuffi2021fixing} that can improve classification performance. 
%%%%
%%%%
\begin{algorithm}[t]
\footnotesize
\SetAlgoLined
\DontPrintSemicolon
\textbf{Input}: Training examples (\textit{X, Y}),  noise bound [-$\epsilon_t$, $\epsilon_t$], noise step $\epsilon_{s}$, learning rate $\eta$, SNN training t-steps $T$, total training epochs $N_{ep}$, iteration $\mathcal{N}$.\\
$\text{// Initialize parameters}$\;
$\kappa \leftarrow 0$ \\
\For{$l \leftarrow 1$ \KwTo $L$}
{
    ${\textbf{W}^l} \leftarrow \text{ANN trained } {\textbf{W}^l}$\;
    ${\textit{v}_{t}^l} \leftarrow \textit{initThreshold} ({\textbf{W}^l}, X)$\;
    ${\textit{l}_{k}^l} \leftarrow  1.0$\;
}
\For{$n \leftarrow 1$ \KwTo $N_{ep}$}
{
%      \uIf{($n \mod I == 0$)}
%      {
%            $\mathcal{N} = \mathcal{N} + 1$\;
%      }
    \For{\text{each batch} $\mathcal{B} \subset (X, Y)$}
    {
        \For{$p \leftarrow 1$ \KwTo ${\mathcal{N}}$}
        {
                $\text{// Compute gradients through STDB}$\;
                ${\delta}_{\textbf{W}} \leftarrow \mathbb{E}_{(x, y) \in \mathcal{B}}[\nabla_{\textbf{W}} \mathcal{L}(g(x + \kappa, y; \frac{T}{\mathcal{N}}))]$\;  
                
                ${\delta}_{\textit{v}_t} \leftarrow \mathbb{E}_{(x, y) \in \mathcal{B}}[\nabla_{\textit{v}_{t}} \mathcal{L}(g(x + \kappa, y;  \frac{T}{\mathcal{N}}))]$\; 
                
                ${\delta}_{\textit{l}_{k}} \leftarrow \mathbb{E}_{(x, y) \in \mathcal{B}}[\nabla_{\textit{l}_{k}} \mathcal{L}(g(x + \kappa, y;  \frac{T}{\mathcal{N}}))]$\; 
                
                $\text{// Compute perturbation}$\;
                $\delta_{x} \leftarrow [\nabla_{x} \mathcal{L}(g(x + \kappa, y; \frac{T}{\mathcal{N}}))]$\;
                $\kappa \leftarrow \text{clip}(\kappa + \epsilon_{s}*sign(\delta_{x}), -\epsilon_t, \epsilon_t)$\;
                
                $\text{// Update trainable parameters}$\;
                ${\textbf{W}} \leftarrow {\textbf{W}} - \eta*{\delta}_{\textbf{W}}$\;
                
                ${\textit{v}_t} \leftarrow {\textit{v}_t} - \eta*{\delta}_{\textit{v}_t}$\;
                
                ${\textit{l}_k} \leftarrow {\textit{l}_k} - \eta*{\delta}_{\textit{l}_k}$\;
            
        }
    }
}
 \caption{HIRE-SNN Training Algorithm}
 \label{alg:scin}
\end{algorithm}
%%%%
%%%%
\section{Experiments}
\label{sec:expt}
%%%%
%%%%
\subsection{Experimental Setup}
\label{subsec:setup}
\textbf{Dataset and ANN training.}
For our experiments we used two widely accepted image classification datasets, namely CIFAR-10 and CIFAR-100. 
For both ANN and direct-input SNN training, we use the standard data-augmented (horizontal flip and random crop with reflective padding) input. For rate-coded input based SNN training, we produce a spike train with rate proportional to the input pixel via a Poisson generator function \cite{Kundu2021WACV}. We performed ANN training for 240 epochs with an initial learning rate (LR) of $0.01$ that decayed by a factor of 0.1 after $150$, $180$, and $210$ epochs.

\textbf{ANN-SNN conversion and SNN training.} We performed the ANN-SNN conversion as recommended in \cite{rathi2020diet} to generate initial thresholds for the SNN training. We then train the converted SNN for only $30$ epochs with batch-size of $32$ starting with the trained ANN weights. We set starting LR to $10^{-4}$ and decay it by a factor of $5$ after $60\%$, $80\%$, and $90\%$ completion of the total training epochs. Unless stated otherwise, we used training time steps $T$ of 6, 8, and 10 for VGG5, VGG11, and ResNet12, respectively. To avoid overfitting and perform regularization we used a dropout of 0.2 to train the models. The $\epsilon_{s}$ is chosen to be $0.013$ and $0.025$ (apart from the $\epsilon_{s}$ sweep test) to train with VGG5 and VGG11, respectively, with $\epsilon_t$ equal to $\epsilon_{s}$. For ResNet12 we chose $\epsilon_{s}$ to be $0.008$ and $0.015$ on CIFAR-10 and CIFAR100, respectively. Also, $\mathcal{N}$ is set to $2$ unless otherwise mentioned. The basic motivation to pick hyperparameters $\mathcal{N}$, $\epsilon_s$, and $\epsilon_t$ is to ensure there is only an insignificant drop in the clean image accuracy while still improving the adversarial performance. We conducted all the experiments on a NVIDIA 2080 Ti GPU having 11 GB memory with the models implemented using PyTorch \cite{paszke2017automatic}. Further training and model details along with analysis on the hyperparameters are provided in the supplementary material.

\textbf{Adversarial test setup.}
For PGD, we set $\epsilon$ for the $L_\infty$ neighborhood to $8/255$, the attack step size  $\alpha = 0.01$, and the number of attack iterations $K$ to $7$, the same values as in \cite{sharmin2020inherent}. For FGSM, we choose the same $\epsilon$ value as above.
%%%%
%%%%
\subsection{Performance Against WB and BB Attacks}
To perform this evaluation, for each model variant we use three differently trained networks: ANN equivalent $\Phi_{ANN}$, hybrid traditionally trained SNN $\Phi_{SNN}^{T}$, and SNN trained with proposed technique $\Phi_{SNN}^{P}$, all trained to have comparable clean-image classification accuracy. We compute $\Delta_d$ as the difference in clean-image classification performance between $\Phi_{SNN}^{P}$ and $\Phi_{SNN}^{T}$. We define $\Delta_a$\footnote{$\Delta_a$ between model M1 and M2 is $Acc_{M1}\% - Acc_{M2}\%$.} as the accuracy difference between $\Phi_{SNN}^{P}$ and either of $\Phi_{SNN}^{T}$ or $\Phi_{ANN}$ while classifying on perturbed image. Note, both $\Phi_{SNN}^{P}$ and $\Phi_{SNN}^{T}$ are trained with direct inputs.
Table \ref{tab:scin_white_box_results} shows the absolute and relative performances of the models generated through our training framework on white-box attack generated images using both FGSM and PGD attack techniques. In particular, 
we observe that with negligible performance compromise on clean images, $\Phi_{SNN}^{P}$ consistently outperforms $\Phi_{SNN}^{T}$ for all the models on both datasets. Specifically, we observe that the perturbed image classification can have an improved performance of up to $12.2\%$ and $8.8\%$, on CIFAR-10 and CIFAR-100 respectively. Compared to $\Phi_{ANN}$ we observe improved performance of up to $25\%$ on WB attacks.
%%%%
%%%%
\begin{table}
\scriptsize\addtolength{\tabcolsep}{-3.5pt}
\begin{center}
{\makegapedcells
\begin{tabular}{c|c|c|c||c|c||c|c}
\hline
{} & \multicolumn{3}{|c||}{Accuracy (\%) with} & \multicolumn{2}{|c||}{$\Delta_a$ over traditional} & \multicolumn{2}{|c}{$\Delta_a$ over ANN} \\
Model & \multicolumn{3}{|c||}{proposed SNN training} & \multicolumn{2}{|c||}{SNN training} & \multicolumn{2}{|c}{ equivalent} \\
\cline{2-8}
{} & Clean($\Delta_d$) & FGSM & PGD  & FGSM & PGD  & FGSM & PGD \\
\hline
 \multicolumn{8}{c}{Dataset : CIFAR-10} \\
\hline
VGG5 & 87.5 (-0.4) & 38.0 & 9.1 & +2.5 & $\textbf{+3.8}$ & $\textbf{+25}$ & $\textbf{+7.1}$ \\
\hline
ResNet12 & 90.3 (-1.6) & 33.3 & 3.8 & $\textbf{+12.2}$ & +3.5 & +13.4 & +1.8 \\
\hline
 \multicolumn{8}{c}{Dataset : CIFAR-100} \\
\hline
VGG11 & 65.1 (-0.4) & 22.0 & 7.5 & +5.7 & +4.6 & +5.1 & -0.7 \\
\hline
ResNet12 & 58.9 (-3.0) & 19.3 & 5.3 & $\textbf{+8.8}$ & $\textbf{+4.7}$ & $\textbf{+5.8}$ & $\textbf{+2.5}$ \\
\hline
\end{tabular}}
\end{center}
\caption{Performance comparison of SNN models generated using the proposed training scheme on clean and adversarially-generated images under a white-box attack.}
\label{tab:scin_white_box_results}
\end{table}

Table \ref{tab:scin_black_box_results} shows the model performances and comparisons on black-box attack generated images using both FGSM and PGD. For this evaluation, for each model variant we used the same model trained with a different seed to generate the perturbed images. For all the models on both the datasets we observe $\Phi_{SNN}^{P}$ yields higher accuracy on the perturbed images generated through BB attack compared to those generated through WB attack, primarily because of BB attacks yield weaker perturbations \cite{athalye2018obfuscated, chen2017zoo}. Importantly, we observe superior performance of $\Phi_{SNN}^{P}$ over both $\Phi_{SNN}^{T}$ and $\Phi_{SNN}^{T}$ under this weaker form of attack. In particular, $\Phi_{SNN}^{P}$ provides an 
improvement $\Delta_a$ of up to $13.7\%$ and $10.4\%$ on CIFAR-10 and CIFAR-100, compared to $\Phi_{SNN}^{T}$. 
%%%%
%%%%
\begin{table}
\scriptsize\addtolength{\tabcolsep}{-3.5pt}
\begin{center}
{\makegapedcells
\begin{tabular}{c|c|c|c||c|c||c|c}
\hline
{} & \multicolumn{3}{|c||}{Accuracy (\%) with} & \multicolumn{2}{|c||}{$\Delta_a$ over traditional} & \multicolumn{2}{|c}{$\Delta_a$ over ANN} \\
Model & \multicolumn{3}{|c||}{proposed SNN training} & \multicolumn{2}{|c||}{SNN training} & \multicolumn{2}{|c}{ equivalent} \\
\cline{2-8}
{} & Clean & FGSM & PGD  & FGSM & PGD  & FGSM & PGD \\
\hline
 \multicolumn{8}{c}{Dataset : CIFAR-10} \\
\hline
VGG5 & 87.5 & 42.1 & 14.9 & +3.9 & $\textbf{+8.3}$ & $\textbf{+18.1}$ & $\textbf{+8.5}$ \\
\hline
ResNet12 & 90.3 & 38.4 & 7.8 & $\textbf{+13.7}$ & +7.2 & +9.7 & +3.5\\
\hline
 \multicolumn{8}{c}{Dataset : CIFAR-100} \\
\hline
VGG11 & 65.1 & 29.1 & 16.1 & +10.0 & +9.9 & $\textbf{+5.6}$ & $\textbf{+0.9}$ \\
\hline
ResNet12 & 58.9 & 24.5 & 12.1 & $\textbf{+10.4}$ & $\textbf{+10.1}$ & +1.3 & $\mathord{\sim}0$ \\
\hline
\end{tabular}}
\end{center}
\caption{Performance comparison of SNN models generated using the proposed training scheme on clean and adversarially-generated images under a black-box attack.}
\vspace{-4mm}
\label{tab:scin_black_box_results}
\end{table}
%%%%
%%%%
\vspace{-4mm}
\begin{figure}[ht]
\includegraphics[width=0.32\textwidth]{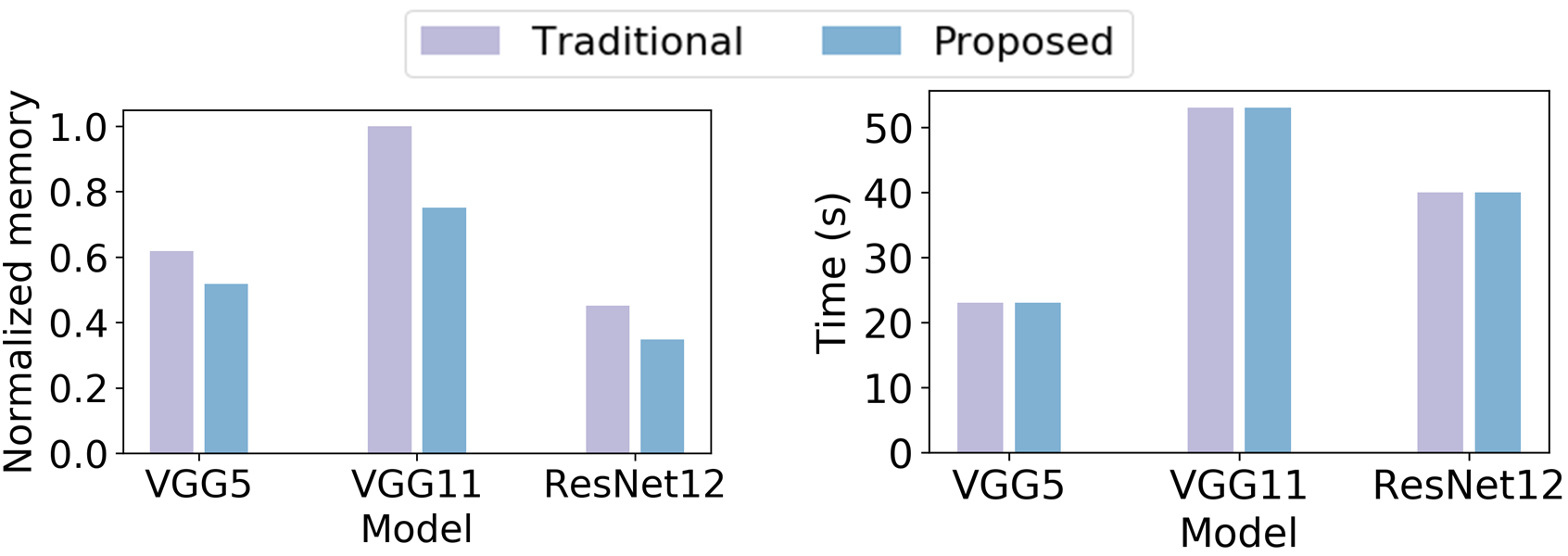}
\centering
   \caption{Normalized GPU memory usage and average training time for a batch of 200 images for VGG5, VGG11, and ResNet12 when trained with the traditional and proposed approaches.}
   %over $T$ time steps.}
\label{fig:ram_and_time_vs_model_comparison}
\vspace{-4mm}
\end{figure}
%%%%

Fig. \ref{fig:ram_and_time_vs_model_comparison} shows the normalized random access memory (RAM) memory and average training time for 200 batches for both the traditional and presented SNN training. Interestingly, due to the shorter update interval the proposed approach require less memory by up to $\mathord{\sim}25\%$ while incurring no extra GPU training time.

\begin{table}[!ht]
\begin{center}
\scriptsize\addtolength{\tabcolsep}{-2.5pt}
\begin{tabular}{lcc}
\hline
Checks to identify gradient obfuscation  & Fail & Pass\\
\hline
\hline
i) Single-step attack performs better compared to iterative attacks &  & \checkmark \\
ii) Black-box attacks performs better compared to white-box attacks &  & \checkmark \\
iii) Increasing perturbation bound can't increase attack strength &  & \checkmark \\
iv) Unbounded attacks can't reach $\mathord{\sim}100\%$ success &   & \checkmark \\
v) Adversarial example can be found through random sampling &  & \checkmark \\
\hline
\end{tabular}
\end{center}
\caption{Checklist set of tests for characteristic behaviors caused by obfuscated and masked gradients \cite{athalye2018obfuscated}.}
\label{tab:check_obfuscation}
\vspace{-2mm}
\end{table}
%%%%
%%%%

\subsection{Discussion}
\label{subsec:disc}
Here, we evaluate the potential presence of obfuscated gradients through experiments with the HIRE-SNN trained models under different attack strengths. We then study the efficacy of noise crafting and performance under no trainable threshold-leak condition. Finally, we evaluate the impact of the new knob $\epsilon_s$ in trading off clean and perturbed image accuracy.

\textbf{Gradient obfuscation analysis.} We conducted several experiments to verify whether the inherent robustness of the presented HIRE-SNNs come from an incorrect approximation of the true gradient based on a single sample. In particular, the performance of generated models was checked against the five tests (Table \ref{tab:check_obfuscation}) proposed in \cite{athalye2018obfuscated} that can identify potential gradient obfuscation. 

\begin{figure}[!ht]
\includegraphics[width=0.48\textwidth]{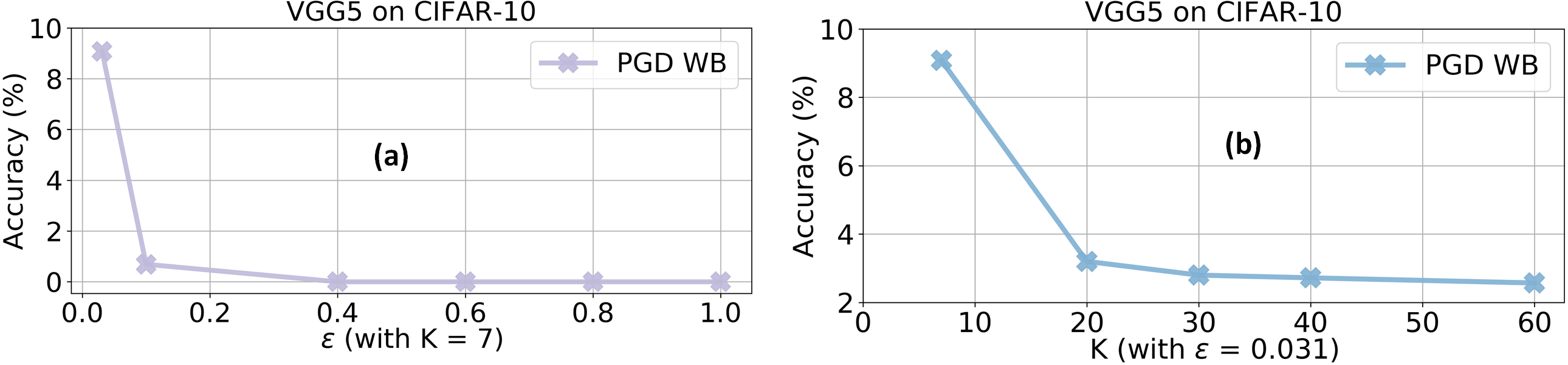}
\centering
   \caption{White-box PGD attack performance as a function of (a) bound $\epsilon$ and (b) attack iterations $K$ with VGG5 on CIFAR-10.}
\label{fig:obfusc_test_vgg5}
\vspace{-2mm}
\end{figure}

As shown in Table \ref{tab:scin_white_box_results} and \ref{tab:scin_black_box_results}, for all the models on both datasets the single-step FGSM performs poorly compared to its iterative counterpart PGD. This certifies the success of Test (i), 
as listed in Table \ref{tab:check_obfuscation}.  
Test (ii) passes because our black-box generated perturbations in Table \ref{tab:scin_black_box_results} yield weaker attacks\footnote{Note that here we say an attack is weaker than other when the classification accuracy on that attack-generated images is higher compared to the images generated through the other.} than their white-box counterparts shown in Table \ref{tab:scin_white_box_results}. To verify Tests (iii) and (iv) we analyzed VGG5 on CIFAR-10 with increasing attack bound $\epsilon$. As shown in Fig. \ref{fig:obfusc_test_vgg5}(a), the classification accuracy decreases as we increase $\epsilon$ and finally reaches an accuracy of $\mathord{\sim}0\%$. %validating both the items. 
Test (v) can fail only if gradient based attacks cannot provide adversarial examples for the model to misclassify. It is clear from our experiments, however, that FGSM and PGD, both variants of gradient based attacks, can sometimes fool the network despite our training. %, thus validates item v) by disproving.

We also evaluated the VGG5 performance with increased attack strength by increasing the number of iterations $K$ of PGD and found that the model's robustness decreases with increasing $K$. However, as Fig. \ref{fig:obfusc_test_vgg5}(b) shows, after $K=40$, the robustness of the model nears an asymptote.
%model retains a minimum robustness which nearly remains constant as $K$ increases further. 
In contrast, if the success of the HIRE-SNNs arose from the incorrect gradient of a single sample, increasing the attack iterations would have broken the defense completely \cite{he2019parametric}. 

Thus, based on these evaluations we conclude that even if the models include obfuscated gradients, they are not significant source of the robustness for the HIRE-SNNs. 

%%%%%%%%%%%%
%%%%%%%%%%%%

\begin{figure}[h]
\vspace{-2mm}
\includegraphics[width=0.22\textwidth]{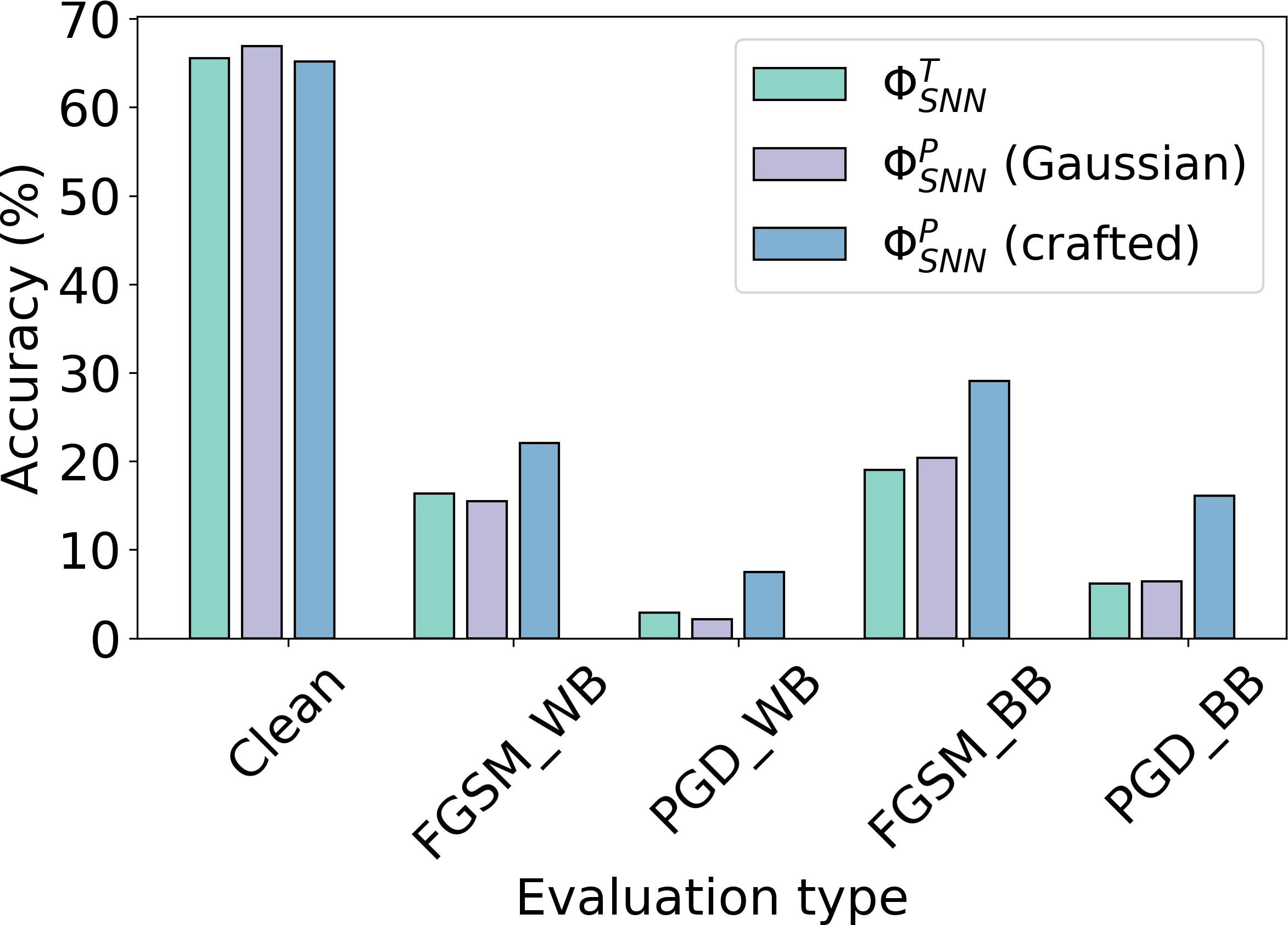}
\centering
   \caption{Comparison of traditional SNN vs. proposed training with both GN and crafted input noise. All the training were performed with direct-input VGG11 on CIFAR-100.}
   %over $T$ time steps.}
\label{fig:vgg11_c100_crafted_vs_gaussian}
\vspace{-2mm}
\end{figure}

\textbf{Importance of careful noise crafting.}
To evaluate the merits of the presented noise crafting technique, we also trained VGG11 with a version of our training algorithm with the perturbation introduced via Gaussian noise. In particular, we pertubed the image pixels using Gaussian noise with zero mean and standard deviation equal to  $\epsilon_s$.
%from which we randomly choose the perturbation to be added to the original pixel values. 
It is clear from Fig.  \ref{fig:vgg11_c100_crafted_vs_gaussian} that compared to the traditional training, the proposed training with perturbation generated through Gaussian noise (GN) fails to provide any noticeable improvement on the adversary-generated images both under white-box and black-box attacks. In contrast, training with carefully crafted noise significantly improves the performance over that with GN against adversary by up to $6.5\%$ and $9.7\%$, on WB and BB attack-created images, respectively.
%%%%%%%%%%%%
%%%%%%%%%%%%
%%%%%%%%%%%%
%%%%%%%%%%%%

\begin{figure*}[!t]
\includegraphics[width=0.94\textwidth]{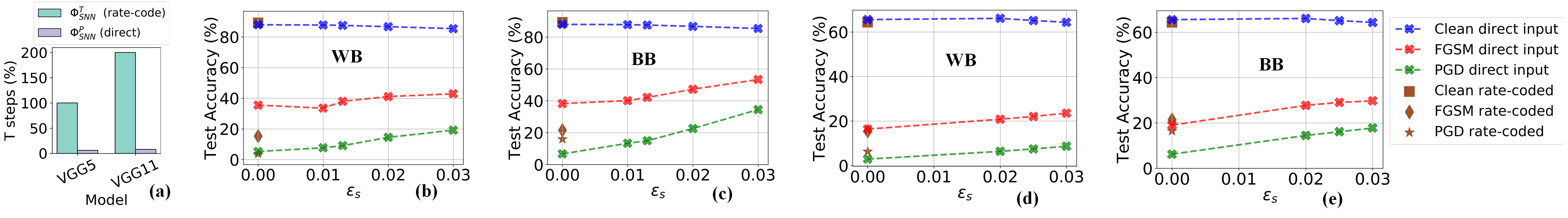}
\centering
   \caption{(a) Inference $T$ steps for rate-coded vs direct input trained SNNs, (b-e) Accuracy vs. $\epsilon_s$ plot for both clean and adversarially generated images (both with WB and BB attack settings) with VGG5 (b, c) and VGG11 (d, e) on CIFAR-10 and CIFAR-100, respectively.}
   %over $T$ time steps.}
\label{fig:vgg5_vgg11_eps_sweep_test}
\vspace{-3mm}
\end{figure*}
%%%%
%%%%
\textbf{Efficacy of proposed training when threshold and leak parameters are not trainable.}  To further evaluate the efficacy of proposed training scheme, we trained VGG5 on CIFAR-10 using our technique but with threshold and leak parameters fixed to their initialized values. As shown in Table \ref{tab:wo_th_lk_comparison}, our generated models still consistently outperform traditionally trained models under both white-box and black-box attacks with negligible drop in clean image accuracy. Interestingly, fixing the threshold and leak parameters yields higher robustness at the cost of lower clean-image accuracy. This may be attributed to the difference in adversarial strength of the perturbed images and is a useful topic of 
future research. 
%We consider evaluation of this to be an interesting future research direction. 

\textbf{Impact of the noise-step knob $\epsilon_{s}$.}
To analyze the impact of the introduced hyperparameter $\epsilon_s$, we performed experiments with VGG5 and VGG11, training the models with various $\epsilon_s \in [0.01, 0.03]$. As depicted in Fig. \ref{fig:vgg5_vgg11_eps_sweep_test} with increased $\epsilon_s$ the models show a consistent improvement on both white-box and black-box attack generated perturbed images with only a small drop in clean image performance of up to $\mathord{\sim}2\%$. Note, here $\epsilon_s = 0$ corresponds to traditional SNN training. With the optimal choice of $\epsilon_s$ our models outperform the state-of-the-art inherently robust SNNs trained on rate-coded inputs \cite{sharmin2020inherent} maintaining similar clean image accuracy with an improved inference latency of up to $25 \times$ as shown in Fig. \ref{fig:vgg5_vgg11_eps_sweep_test}. 
%These results clearly show the success of our training scheme to harness the inherent robustness.  
%%%%
%%%%
\begin{table}
\scriptsize\addtolength{\tabcolsep}{-4.0pt}
\begin{center}
{\makegapedcells
\begin{tabular}{c|c|c|c||c|c||c|c}
\hline
 Model & Dataset & Training & Clean & \multicolumn{2}{|c||}{Acc. $\%$ on WB} & \multicolumn{2}{|c}{Acc. $\%$ on BB} \\
\cline{5-8}
 {} & {} & Method & Acc. ($\%$) & FGSM & PGD  & FGSM & PGD \\
\hline
 VGG5 & CIFAR-10 & Traditional & $\textbf{87.2}$ & 33.0 & 4.5 & 40.4 & 8.8 \\
 \cline{3-8}
  {} & {} & Proposed & 86.8 & $\textbf{40.5}$ & $\textbf{13.6}$ & $\textbf{46.2}$ & $\textbf{21.9}$ \\
\hline
\end{tabular}}
\end{center}
\caption{Performance comparison of proposed with traditional SNN training when threshold-leak parameters are frozen to their initialized values.}
\vspace{-5mm}
\label{tab:wo_th_lk_comparison}
\end{table}
%%%%
%%%%
\subsection{Computation Energy}
\label{subsec:compute_ener}

\begin{table}[!ht]
\vspace{-5mm}
\scriptsize\addtolength{\tabcolsep}{-2pt}
\begin{center}
\begin{tabular}{c|c|c}
\hline
{Model} & \multicolumn{2}{|c}{FLOPs of a CONV layer $l$}\\
\cline{2-3}
        & Variable & Value \\
\hline
{$ANN$} \cite{kundu2020pre} &  $FL_{ANN}^l$ & $(k^l)^2\times H_o^l\times W_o^l\times C_o^l\times C_i^l$\\
\hline
{$SNN$} \cite{Kundu2021WACV} & $FL_{SNN}^l$ & $(k^l)^2\times H_o^l\times W_o^l\times C_o^l\times C_i^l \times {\zeta}^l$ \\
\hline
\end{tabular}
\end{center}
\caption{Convolutional layer FLOPs for ANN and SNN models}
\vspace{-3mm}
\label{tab:flops}
\end{table}

\begin{figure}[ht]
\includegraphics[width=0.38\textwidth]{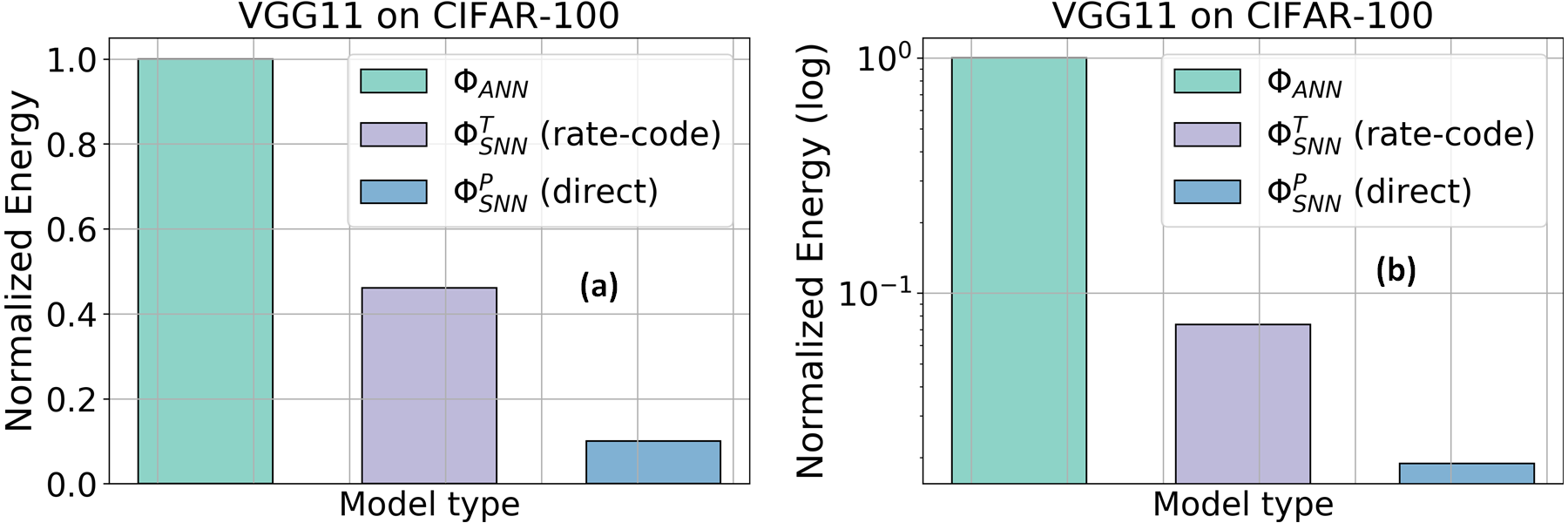}
\centering
   \caption{Comparison of normalized compute energy computed assuming (a) 32-bit FP and (b) 32-bit INT implementations.}
   %over $T$ time steps.}
\label{fig:ec_comparsion_float_fixed}
\vspace{-2mm}
\end{figure}

Let us assume a convolutional layer $l$ with weight tensor $\textbf{W}^l \in \mathbb{R}^{k^l \times k^l \times C_{i}^l \times C_{o}^l}$ taking an input activation tensor $\textbf{A}^l \in \mathbb{R}^{H_{i}^l \times W_{i}^l \times C_{i}^l}$, with $ H_{i}^l, W_{i}^l$, $k^l$, $C_i^l$ and $C_{o}^l$ to be the input height, width, filter height (and width), channel size, and number of filters, respectively. Table \ref{tab:flops} presents the FLOPs requirement for an ANN and corresponding SNN for this layer to produce an output activation tensor $\textbf{O}^l \in \mathbb{R}^{H_{o}^l \times W_{o}^l \times C_{o}^l}$. ${\zeta}^l$ represents the associated spiking activity for layer $l$. Now, for an $L$-layer SNN with rate-coded and direct inputs, the inference computation energy is,
\begin{align} 
\footnotesize
& E_{SNN}^{rate}=(\sum_{l=1}^{L}FL^l_{SNN})\cdot{E_{AC}} \\
& E_{SNN}^{direct}= FL_{SNN}^1\cdot{E_{MAC}} + (\sum_{l=2}^{L}FL^l_{SNN})\cdot{E_{AC}} 
\end{align}
\noindent
where $E_{AC}$ and $E_{MAC}$ represent the energy cost of AC and MAC operation, respectively. For our evaluation we use their values as shown in Table \ref{tab:fp_int_energy}. In particular, as exemplified in Fig. \ref{fig:ec_comparsion_float_fixed}(a), the computation energy benefit of HIRE-SNN VGG11 over its inherently robust rate-coded SNN and ANN 
counterpart is as high as $4.6 \times$ and $10 \times$, respectively, considering 32-b floating point (FP) representation. For a 32-b integer (INT) implementation, this advantage is as much as $3.9\times$ and $53 \times$, respectively (Fig. \ref{fig:ec_comparsion_float_fixed}(b)).
%%%%%%%%%%%%
%%%%%%%%%%%%
%%%%%%%%%%%%
%%%%%%%%%%%%

\begin{table}[!ht]
\scriptsize\addtolength{\tabcolsep}{-2.5pt}
\begin{center}
\begin{tabular}{c|c|c|c}
\hline
{Serial} & {Operation} & \multicolumn{2}{c}{Energy ($pJ$)} \\
\cline{3-4}
{ No.} & {}  &  32-b INT & 32-b FP \\
\hline
1. & 32-bit multiplication & $3.1$ & $3.7$ \\
2. & 32-bit addition & $0.1$ & $0.9$ \\
\hline
3. & 32-bit MAC ($\#1$ + $\#2$) &  $3.2$ & $4.6$ \\
4. & 32-bit AC ($\#2$) & $0.1$ & $0.9$\\
\hline
\end{tabular}
\end{center}
\caption{Estimated energy costs for various operations in a 45 $nm$ CMOS process at 0.9 V \cite{horowitz20141}}
\vspace{-5mm}
\label{tab:fp_int_energy}
\end{table}
%%%%%%%%%%%%
%%%%%%%%%%%%
\section{Conclusions}
\label{sec:conc}
In this paper we first analyzed the inherent robustness of low-latency SNNs trained with direct inputs to provide insightful observations. Motivated by these observations we then present a training algorithm that harnesses the inherent robustness of low-latency SNNs without incurring any additional training time cost. We conducted extensive experimental analysis to evaluate the efficacy of our training along with experiments to understand the contribution of the carefully crafted noise. Particularly, compared to traditionally trained direct input SNNs, the generated SNNs can yield accuracy improvement of up to $13.7\%$ on black-box FGSM attack generated images. Compared to the SOTA inherently robust VGG11 SNN trained on rate-coded inputs (CIFAR-100) our models perform similarly or better on clean and perturbed image classification performance while providing an improved performance of up to $25\times$ and $\mathord{\sim}4.6\times$, in terms of inference latency and computation energy, respectively. We believe that this study is a step in making deep SNNs a practical energy-efficient solution for safety-critical inference applications where robustness is a need.

\section{Acknowledgments}
\label{sec:ack}
This work was  partly supported by USC Annenberg fellowship and  NSF including grant number $1763747$. 
%-------------------------------------------------------------------------

%------------------------------------------------------------------------

%-------------------------------------------------------------------------

%----------------------------------------------------

%-------------------------------------------------------------------------

%-------------------------------------------------------------------------

%-------------------------------------------------------------------------

%-------------------------------------------------------------------------

%------------------------------------------------------------------------

{\small
\bibliographystyle{ieee_fullname}
\bibliography{egbib}

\begin{thebibliography}{10}\itemsep=-1pt

\bibitem{athalye2018obfuscated}
Anish Athalye, Nicholas Carlini, and David Wagner.
\newblock Obfuscated gradients give a false sense of security: Circumventing
  defenses to adversarial examples.
\newblock In {\em International Conference on Machine Learning}, pages
  274--283. PMLR, 2018.

\bibitem{bellec_2018long}
Guillaume Bellec, Darjan Salaj, Anand Subramoney, Robert Legenstein, and
  Wolfgang Maass.
\newblock Long short-term memory and learning-to-learn in networks of spiking
  neurons.
\newblock {\em arXiv preprint arXiv:1803.09574}, 2018.

\bibitem{chen2017zoo}
Pin-Yu Chen, Huan Zhang, Yash Sharma, Jinfeng Yi, and Cho-Jui Hsieh.
\newblock Zoo: Zeroth order optimization based black-box attacks to deep neural
  networks without training substitute models.
\newblock In {\em Proceedings of the 10th ACM workshop on artificial
  intelligence and security}, pages 15--26, 2017.

\bibitem{datta2021training}
Gourav Datta, Souvik Kundu, and Peter~A Beerel.
\newblock Training energy-efficient deep spiking neural networks with
  single-spike hybrid input encoding.
\newblock {\em arXiv preprint arXiv:2107.12374}, 2021.

\bibitem{deng2012mnist}
Li Deng.
\newblock The {MNIST} database of handwritten digit images for machine learning
  research [best of the web].
\newblock {\em IEEE Signal Processing Magazine}, 29(6):141--142, 2012.

\bibitem{dsnn_conversion_ijcnn}
P.~U. {Diehl}, D. {Neil}, J. {Binas}, M. {Cook}, S. {Liu}, and M. {Pfeiffer}.
\newblock Fast-classifying, high-accuracy spiking deep networks through weight
  and threshold balancing.
\newblock In {\em 2015 International Joint Conference on Neural Networks
  (IJCNN)}, volume~1, pages 1--8, 2015.

\bibitem{el2020securing}
Rida El-Allami, Alberto Marchisio, Muhammad Shafique, and Ihsen Alouani.
\newblock Securing deep spiking neural networks against adversarial attacks
  through inherent structural parameters.
\newblock {\em arXiv preprint arXiv:2012.05321}, 2020.

\bibitem{farabet2012comparison}
Cl{\'e}ment Farabet, Rafael Paz, Jose P{\'e}rez-Carrasco, Carlos Zamarre{\~n}o,
  Alejandro Linares-Barranco, Yann LeCun, Eugenio Culurciello, Teresa
  Serrano-Gotarredona, and Bernabe Linares-Barranco.
\newblock Comparison between frame-constrained fix-pixel-value and frame-free
  spiking-dynamic-pixel convnets for visual processing.
\newblock {\em Frontiers in neuroscience}, 6:32, 2012.

\bibitem{goodfellow2014explaining}
Ian~J. Goodfellow, Jonathon Shlens, and Christian Szegedy.
\newblock Explaining and harnessing adversarial examples.
\newblock {\em arXiv preprint arXiv:1412.6572}, 2014.

\bibitem{he2016deep}
Kaiming He, Xiangyu Zhang, Shaoqing Ren, and Jian Sun.
\newblock Deep residual learning for image recognition.
\newblock In {\em Proceedings of the IEEE conference on computer vision and
  pattern recognition}, pages 770--778, 2016.

\bibitem{he2019parametric}
Zhezhi He, Adnan~Siraj Rakin, and Deliang Fan.
\newblock Parametric noise injection: Trainable randomness to improve deep
  neural network robustness against adversarial attack.
\newblock In {\em Proceedings of the IEEE/CVF Conference on Computer Vision and
  Pattern Recognition}, pages 588--597, 2019.

\bibitem{horowitz20141}
Mark Horowitz.
\newblock 1.1 {Computing's} energy problem (and what we can do about it).
\newblock In {\em 2014 IEEE International Solid-State Circuits Conference
  Digest of Technical Papers (ISSCC)}, pages 10--14. IEEE, 2014.

\bibitem{neuro_frontiers}
Giacomo Indiveri and Timothy Horiuchi.
\newblock Frontiers in neuromorphic engineering.
\newblock {\em Frontiers in Neuroscience}, 5:118, 2011.

\bibitem{krizhevsky2009learning}
Alex Krizhevsky, Geoffrey Hinton, et~al.
\newblock Learning multiple layers of features from tiny images.
\newblock 2009.

\bibitem{Kundu2021WACV}
Souvik Kundu, Gourav Datta, Massoud Pedram, and Peter~A. Beerel.
\newblock Spike-thrift: Towards energy-efficient deep spiking neural networks
  by limiting spiking activity via attention-guided compression.
\newblock In {\em Proceedings of the IEEE/CVF Winter Conference on Applications
  of Computer Vision (WACV)}, pages 3953--3962, January 2021.

\bibitem{kundu2021towards}
Souvik Kundu, Gourav Datta, Massoud Pedram, and Peter~A Beerel.
\newblock Towards low-latency energy-efficient deep snns via attention-guided
  compression.
\newblock {\em arXiv preprint arXiv:2107.12445}, 2021.

\bibitem{kundu2021dnr}
Souvik Kundu, Mahdi Nazemi, Peter~A Beerel, and Massoud Pedram.
\newblock Dnr: A tunable robust pruning framework through dynamic network
  rewiring of dnns.
\newblock In {\em Proceedings of the 26th Asia and South Pacific Design
  Automation Conference}, pages 344--350, 2021.

\bibitem{kundu2020pre}
Souvik Kundu, Mahdi Nazemi, Massoud Pedram, Keith~M Chugg, and Peter Beerel.
\newblock Pre-defined sparsity for low-complexity convolutional neural
  networks.
\newblock {\em IEEE Transactions on Computers}, 2020.

\bibitem{kundu2021attentionlite}
Souvik Kundu and Sairam Sundaresan.
\newblock Attentionlite: Towards efficient self-attention models for vision.
\newblock In {\em ICASSP 2021-2021 IEEE International Conference on Acoustics,
  Speech and Signal Processing (ICASSP)}, pages 2225--2229. IEEE, 2021.

\bibitem{leefin2020}
Chankyu Lee, Syed~Shakib Sarwar, Priyadarshini Panda, Gopalakrishnan
  Srinivasan, and Kaushik Roy.
\newblock Enabling spike-based backpropagation for training deep neural network
  architectures.
\newblock {\em Frontiers in Neuroscience}, 14:119, 2020.

\bibitem{lu2020exploring}
Sen Lu and Abhronil Sengupta.
\newblock Exploring the connection between binary and spiking neural networks.
\newblock {\em arXiv preprint arXiv:2002.10064}, 2020.

\bibitem{madry2017towards}
Aleksander Madry, Aleksandar Makelov, Ludwig Schmidt, Dimitris Tsipras, and
  Adrian Vladu.
\newblock Towards deep learning models resistant to adversarial attacks.
\newblock {\em arXiv preprint arXiv:1706.06083}, 2017.

\bibitem{mainen1995reliability}
Zachary~F Mainen and Terrence~J Sejnowski.
\newblock Reliability of spike timing in neocortical neurons.
\newblock {\em Science}, 268(5216):1503--1506, 1995.

\bibitem{marchisio2020spiking}
Alberto Marchisio, Giorgio Nanfa, Faiq Khalid, Muhammad~Abdullah Hanif,
  Maurizio Martina, and Muhammad Shafique.
\newblock Is spiking secure? {A} comparative study on the security
  vulnerabilities of spiking and deep neural networks.
\newblock In {\em 2020 International Joint Conference on Neural Networks
  (IJCNN)}, pages 1--8. IEEE, 2020.

\bibitem{moosavi2016deepfool}
Seyed-Mohsen Moosavi-Dezfooli, Alhussein Fawzi, and Pascal Frossard.
\newblock {DeepFool}: a simple and accurate method to fool deep neural
  networks.
\newblock In {\em Proceedings of the IEEE conference on computer vision and
  pattern recognition}, pages 2574--2582, 2016.

\bibitem{neftci_surg}
E.~O. {Neftci}, H. {Mostafa}, and F. {Zenke}.
\newblock Surrogate gradient learning in spiking neural networks: Bringing the
  power of gradient-based optimization to spiking neural networks.
\newblock {\em IEEE Signal Processing Magazine}, 36(6):51--63, 2019.

\bibitem{paszke2017automatic}
Adam Paszke, Sam Gross, Soumith Chintala, Gregory Chanan, Edward Yang, Zachary
  DeVito, Zeming Lin, Alban Desmaison, Luca Antiga, and Adam Lerer.
\newblock Automatic differentiation in pytorch.
\newblock 2017.

\bibitem{spike_ratecoding}
Michael Pfeiffer and Thomas Pfeil.
\newblock Deep learning with spiking neurons: Opportunities and challenges.
\newblock {\em Frontiers in Neuroscience}, 12:774, 2018.

\bibitem{rathi2020diet}
Nitin Rathi and Kaushik Roy.
\newblock {DIET-SNN}: Direct input encoding with leakage and threshold
  optimization in deep spiking neural networks.
\newblock {\em arXiv preprint arXiv:2008.03658}, 2020.

\bibitem{rebuffi2021fixing}
Sylvestre-Alvise Rebuffi, Sven Gowal, Dan~A Calian, Florian Stimberg, Olivia
  Wiles, and Timothy Mann.
\newblock Fixing data augmentation to improve adversarial robustness.
\newblock {\em arXiv preprint arXiv:2103.01946}, 2021.

\bibitem{redmon2017yolo9000}
Joseph Redmon and Ali Farhadi.
\newblock {YOLO9000}: better, faster, stronger.
\newblock In {\em Proceedings of the IEEE conference on computer vision and
  pattern recognition}, pages 7263--7271, 2017.

\bibitem{dsnn_conversion_abhronilfin}
Abhronil Sengupta, Yuting Ye, Robert Wang, Chiao Liu, and Kaushik Roy.
\newblock Going deeper in spiking neural networks: {VGG} and residual
  architectures.
\newblock {\em Frontiers in Neuroscience}, 13:95, 2019.

\bibitem{shafahi2019adversarial}
Ali Shafahi, Mahyar Najibi, Amin Ghiasi, Zheng Xu, John Dickerson, Christoph
  Studer, Larry~S Davis, Gavin Taylor, and Tom Goldstein.
\newblock Adversarial training for free!
\newblock {\em arXiv preprint arXiv:1904.12843}, 2019.

\bibitem{sharmin2020inherent}
Saima Sharmin, Nitin Rathi, Priyadarshini Panda, and Kaushik Roy.
\newblock Inherent adversarial robustness of deep spiking neural networks:
  Effects of discrete input encoding and non-linear activations.
\newblock In {\em European Conference on Computer Vision}, pages 399--414.
  Springer, 2020.

\bibitem{simonyan2014very}
Karen Simonyan and Andrew Zisserman.
\newblock Very deep convolutional networks for large-scale image recognition.
\newblock {\em arXiv preprint arXiv:1409.1556}, 2014.

\bibitem{szegedy2015going}
Christian Szegedy, Wei Liu, Yangqing Jia, Pierre Sermanet, Scott Reed, Dragomir
  Anguelov, Dumitru Erhan, Vincent Vanhoucke, and Andrew Rabinovich.
\newblock Going deeper with convolutions.
\newblock In {\em Proceedings of the IEEE Conference on Computer Vision and
  Pattern Recognition}, pages 1--9, 2015.

\bibitem{tao2018image}
Hu Tao, Weihua Li, Xianxiang Qin, and Dan Jia.
\newblock Image semantic segmentation based on convolutional neural network and
  conditional random field.
\newblock In {\em 2018 Tenth International Conference on Advanced Computational
  Intelligence (ICACI)}, pages 568--572. IEEE, 2018.

\bibitem{wu2018spatio}
Yujie Wu, Lei Deng, Guoqi Li, Jun Zhu, and Luping Shi.
\newblock Spatio-temporal backpropagation for training high-performance spiking
  neural networks.
\newblock {\em Frontiers in neuroscience}, 12:331, 2018.

\bibitem{xie2017adversarial}
Cihang Xie, Jianyu Wang, Zhishuai Zhang, Yuyin Zhou, Lingxi Xie, and Alan
  Yuille.
\newblock Adversarial examples for semantic segmentation and object detection.
\newblock In {\em Proceedings of the IEEE International Conference on Computer
  Vision}, pages 1369--1378, 2017.

\bibitem{zhang2019you}
Dinghuai Zhang, Tianyuan Zhang, Yiping Lu, Zhanxing Zhu, and Bin Dong.
\newblock You only propagate once: Accelerating adversarial training via
  maximal principle.
\newblock {\em arXiv preprint arXiv:1905.00877}, 2019.

\end{thebibliography}
}

\end{document}